\documentclass[12pt,journal,compsoc,onecolumn]{IEEEtran}
\usepackage{amsmath,epsfig}
 \usepackage{lingmacros}
 \usepackage{tree-dvips}
\usepackage{amsmath} 
\usepackage{amssymb}
\usepackage{pifont}
\usepackage[noadjust]{cite}
\newcommand{\cmark}{\ding{51}}
\newcommand{\xmark}{\ding{55}}

\newlength{\bibitemsep}
\newlength{\bibparskip}
\let\oldthebibliography\thebibliography
\renewcommand\thebibliography[1]{
  \oldthebibliography{#1}
  \setlength{\parskip}{\bibitemsep}
  \setlength{\itemsep}{\bibparskip}
}

\def\x{{\mathbf x}}

\def\XX{{\bf X}}

\def\Y{{\cal Y}}
\def\S{{\cal I}}

\def\x{{\bf x}}
\def\y{{\bf y}}
\def\D{{\cal D}}

\def\F{{ f}}
\def\DD{{d}}
\def\tr{{\bf tr}}
\def\1{{\bf 1}}
\def\I{{\cal I}} 
\def\p{{p}}
\def\q{{q}}
\def\x{{\bf x}}  
\def\Y{{\cal  Y}}  
\def\V{{\bf D}}
\newtheorem{proposition}{Proposition}

\usepackage[ruled,vlined]{algorithm2e}

\title{Adversarial Virtual Exemplar Learning for Label-Frugal Satellite Image Change Detection} 
\author{Hichem Sahbi$^1$ \ \ \ \ \ \  \ \ \ \ \ \  \ \ \ \ \ \   \ \ \ \ \ \ Sebastien Deschamps$^{1,2}$ \\
$ $ \\
$^1$ Sorbonne University, CNRS, LIP6,  F-75005, Paris, France  \\
$^2$Theresis Thales, France
}

\begin{document}
 \maketitle
\begin{abstract}

Satellite image change detection aims at finding occurrences of targeted changes in a given scene taken at different instants.  This task is highly challenging due to the acquisition conditions and also to the subjectivity of changes.  In this paper, we investigate satellite image change detection using active learning. Our method is interactive and  relies on a question \& answer model which asks the oracle (user) questions about  the most informative display (dubbed as virtual exemplars), and according to the user's responses,  updates change detections.   The main contribution of our method consists  in a novel adversarial model that allows  frugally probing the oracle  with only the most representative, diverse and uncertain virtual exemplars.  The latter  are learned to challenge the most the trained change decision criteria which ultimately leads to a better re-estimate of these criteria in the following iterations of active learning.  Conducted  experiments show the out-performance of our proposed adversarial display model against other display strategies as well as the related work.   
\end{abstract}

\section{Introduction}
\label{sec:intro}
Automatic satellite image change detection  consists in finding  occurrences of {\it relevant}  changes  in a given area  at an instant $t_1$ w.r.t.  the same area at an earlier instant $t_0$.  This task is   useful in different applications including remote sensing damage assessment  after natural  hazards (tornadoes, flash-floods, etc.) in order to prioritize  disaster response and rescues   \cite{ref4,ref5}.  Change detection  is  also known to be highly challenging as observed  scenes  are   subject to many irrelevant changes due to  sensors,  occlusions,   radiometric variations and shadows,   weather conditions as well as scene content.   Early change detection solutions  were  based on simple comparisons of multi-temporal signals, via image differences and thresholding,    vegetation indices, principal component and change vector analyses \cite{ref7,ref9,ref11,ref13}.   Other solutions require preliminary preprocessing techniques that mitigate the effect of irrelevant changes by correcting radiometric variations, occlusions, and by estimating the parameters of sensors for registration,  etc. \cite{ref14,ref15,ref17,ref20,refffabc8}  or consider these irrelevant changes as a part of statistical appearance modeling \cite{ref21,ref25,ref26,ref27,ref28,refffabc5,refffabc6,refffabc7}. \\
\indent Among the existing change detection methods, those based on statistical machine learning are particularly effective. However, the success of these methods is bound to the availability of large labeled training sets that  comprehensively capture the huge  variability in  irrelevant changes as well as the user's targeted relevant changes~\cite{refffabc0,refffabc1}.  In practice,   labeled sets are scarce and even when available, their labeling may not reflect the user's subjectivity and intention.  Several alternative solutions seek to make machine learning methods frugal and less (labeled) data-hungry \cite{refffabc3,refffabc4} including few shot \cite{reff45} and self-supervised  learning \cite{refff2}; however,   these methods are agnostic  to the users' intention.  Other solutions,   based on active learning  \cite{reff1,reff2,reff16,reff15,reff53,reff12,reff74,reff58,reff13},  are rather more suitable where users label very few examples of relevant and irrelevant changes according to their intention,  prior to train and retrain  user-dedicated  change detection criteria. \\
\indent In this paper, we devise a novel iterative satellite image change detection solution based on a question \& answer model  that frugally queries the  intention of the user (oracle), and updates change detection results  accordingly.  These queries are restricted {\it only} to the most informative subset of exemplars (also referred to as virtual displays) which  are {\it learned} instead of being sampled from the {\it fixed} pool of unlabeled data.   The informativeness of exemplars is modeled with a  conditional probability distribution that measures how relevant is a given exemplar in the learned displays given the pool of unlabeled  data.  Two novel adversarial losses are also introduced  in order  to learn the display  ---  with the most diverse, representative,  and ambiguous exemplars --- that  challenge the most the previously trained change detection criteria resulting into a better re-estimation of these criteria at the subsequent iterations of change detection. 
While being adversarial,  the two proposed losses are conceptually different from generative adversarial networks (GANs) \cite{refffabc};  GANs aim at producing fake data that mislead the trained  discriminators whilst our proposed adversarial losses seek to generate the most informative data for further annotations. In other words,  the proposed framework allows to sparingly probe the oracle only on the most representative, diverse and uncertain exemplars  that challenge the most the current discriminator, and eventually lead to more accurate ones in the following iterations of change detection.   Achieved experiments corroborate these findings and show the effectiveness of our {\it exemplar and display learning}  models against comparative methods.

\section{Proposed method}
\label{sec:format}
Let's consider ${\cal I}=\{\x_i=(\p_i,\q_i)\}_{i=1}^n \subset \mathbb{R}^d$ as a collection of aligned patch pairs taken from two satellite images captured at two different instants $t_0$, $t_1$, and let $\Y = \{\y_1, \dots, \y_n\} \subset \{-1,+1\}$  be the underlying unknown labels. Our goal is to train a classifier $f: {\cal I} \rightarrow \{-1,+1\}$ that predicts the unknown labels in $\{\y_i\}_i$ with $\y_i = +1$ if the patch $\q_i$  corresponds to a ``change'' w.r.t. the underlying patch $\p_i$, and $\y_i = -1$ otherwise. Training $f$ requires a subset of hand-labeled data obtained from an oracle.  As obtaining these labels is usually highly expensive, the design of $f$ should be {\it label-frugal} while being as accurate  as possible. 
\subsection{Interactive change detection}
Our change detection algorithm is built upon a question \& answer iterative process which consists in (i) submitting the {\it most informative patch pairs} to query their labels from an oracle,  and (ii) updating a classifier $f$ accordingly. In the sequel of this paper, the subset of informative images is dubbed as {\it display}. Let $\D_t \subset \I$ be a display shown to the oracle at iteration $t$, and $\Y_t$ as the unknown labels of $\D_t$; considering a random display $\D_0$, we train our change detection criteria iteratively (for $t \in \{0,\dots,T-1\}$) according to the subsequent steps\\

\noindent 1/ Probe the oracle with  $\D_t$ to obtain $\Y_t$,  and train a decision function $f_t (.)$ on $\cup_{k=0}^t (\D_k,\Y_k)$.; the latter corresponds to a support vector machine (SVM) trained on top of  graph convolution network features \cite{sahbi2021b}.\\
2/ Select the subsequent display $\D_{t+1}\subset \S\backslash\cup_{k=0}^t \D_k$. A strategy that {\it bruteforces} all the possible displays $\D \subset \S\backslash\cup_{k=0}^t \D_k$, trains the associated classifiers  $f_{t+1} (.)$ on $\D \cup_{k=0}^t \D_t$,  and maintains   the display  with the highest accuracy is combinatorial and clearly intractable. Besides, collecting labels on each of these displays is also out of reach. In this work, we consider instead display selection strategies, based on active learning, which are rather more tractable; nonetheless, these display strategies should be carefully designed as many of them are equivalent to (or worse than) basic display strategies that select data uniformly randomly (see for instance \cite{reff2} and references therein). \\
Our proposed display model (as the main contribution of this paper) is different from usual sampling strategies (see e.g.  Sahbi et al.,  IGARSS 2021) and relies on synthesizing exemplars, also referred to as virtual exemplars. The latter provide more flexibility (in exploring the uncharted parts of the space of unlabeled data), and these exemplars are found by maximizing diversity, representativity and uncertainty (as shown in the remainder of this paper). Diversity seeks to hallucinate exemplars that allow exploring uncharted parts of $f_{t+1} (.)$ whereas representativity makes it possible to register the  exemplars, as much as possible, to the input data. Finally, ambiguity locally refines the boundaries of the trained classifiers $f_{t+1} (.)$.  

\subsection{Virtual exemplar learning: early model}
\noindent   We consider for each sample $\x_i \in \I$ a membership distribution $\{\mu_{ik}\}_{k=1}^K$ that measures the {\it conditional} probability of assigning $\x_i$ to K-virtual exemplars; the latter constitute the subsequent display  $\D_{t+1}$.  The proposed method defines $\D_{t+1}$ (rewritten for short and in a matrix form as $\bf D$) together with $\{\mu_{ik}\}_{i,k}$ by minimizing the following constrained objective function  
{ 
\begin{equation}\label{eq0}
\begin{array}{ll}
\displaystyle   \min_{\V; \mu \in \Omega}    & \displaystyle   \tr\big (\mu \ \DD(\V,\XX) \big)  \  + \ \alpha  \ \bigg[\frac{1}{n} \1^\top_n  \mu\bigg] \log \bigg[\frac{1}{n}\1^\top_n \mu\bigg]^\top \\
                 &  \  +  \  \beta \ \tr\big(\F(\V)^\top \  \log \F(\V)\big) \ + \ \gamma \ \tr(\mu^\top \log \mu), 
\end{array}
\end{equation}}
\noindent  here  $\Omega=\{\mu :  \mu \geq 0; \mu \1_K = \1_n\}$,  $\1_{K}$, $\1_{n}$ denote two vectors of $K$ and $n$ ones respectively,  $^\top$ is the matrix transpose operator,  $\mu \in \mathbb{R}^{n \times K}$ is a learned matrix whose i-th row corresponds to the conditional probability of assigning $\x_i$ to each of the K-virtual exemplars, and $\log$ is applied entrywise. In Eq.~\ref{eq0}, $\DD(\V,\XX) \in \mathbb{R}^{K \times n}$ is the matrix of the euclidean distances between input data in $\XX$ and the virtual exemplars in $\V$ whereas $\F(\V) \in \mathbb{R}^{2 \times K} $ is a scoring matrix whose columns correspond to the softmax  of the learned SVM $f_t(.)$ (and its complement) applied to the K-virtual exemplars. The first term in Eq.~\ref{eq0}, rewritten as $\sum_i \sum_k \mu_{ik}  \| \x_i - \V_{k}\|_2^2$, measures how representative are  the virtual examplars $\{\V_{k}\}_k$ w.r.t. the training samples $\XX$. The second term (equal to  $\sum_{k}   [\frac{1}{n} \sum_{i=1}^n  \mu_{ik}]  \log  [\frac{1}{n} \sum_{i=1}^n  \mu_{ik}]$) captures the {\it diversity} of the generated virtual data as the entropy of the probability distribution of the underlying memberships; this measure is minimal when  input data are assigned to different virtual exemplars, and vice-versa. The third criterion (equivalent to $\sum_k \sum_c [\F_c(\V)]^\top_k [\log \F_c(\V)]_k$) measures the {\it ambiguity} (or uncertainty) in $\D_{t+1}$ as the entropy of the scoring function; it reaches its smallest value when virtual exemplars in  $\D_{t+1}$ are evenly scored w.r.t different classes. The fourth term acts as a regularizer which considers that, without any a priori on the three other terms, the conditional probability distribution $\{\mu_{ik}\}_k$ on each input data is uniform. As shown subsequently, this term also helps obtaining a closed form solution. Finally, we consider equality and inequality constraints which guarantee that the membership of each input sample $\x_i$ to the virtual exemplars forms a conditional probability distribution.   

\subsection{Virtual exemplar learning:  surrogate model} The formulation proposed in the aforementioned section, in spite of being relatively effective (as shown later in experiments),  has a downside: it combines heteregeneous terms (entropy and distance based criteria) whose mixing hyperparameters are difficult to optimize\footnote{This requires  trying many settings of these hyperparameters, inducing the underlying displays and labeling them by the oracle. This is clearly out of reach in interactive satellite image change detection as labeling changes/no-changes should be very sparingly and frugally achieved.}. In what follows,  we consider a  surrogate objective function which considers only homogeneous (distance based) criteria, excepting the regularizer,  and this turns out to be more effective as shown later in experiments.  With this variant, the virtual exemplars together with their distributions  $\{\mu_{ik}\}_{i,k}$ are now obtained by minimizing  the following   problem 
{ 
\begin{equation}\label{eq01}
\hspace{-0.25cm}\begin{array}{ll}
\displaystyle   \min_{\V; \mu \in \Omega}    &     \tr\big (\mu \ \DD(\V,\XX)   \big)  \  + \ \frac{\alpha}{2} \  \big\|  \frac{1}{n}\1_n^\top \mu  -\frac{1}{K} \1_K^\top    \big\|_F^2 \\
                 &  \  +  \  \frac{\beta}{2} \   \big\| \F(\V)  - \frac{1}{2} \1_2 \1_K^\top   \big\|_F^2     \ + \ \gamma \ \tr(\mu^\top \log \mu),
\end{array}
\end{equation}}
\noindent here $\|. \|_F$  is the Frobenius norm. The first and fourth terms of the above objective function remain unchanged while the second and third terms,  despite being different,   {\it  their impact is strictly equivalent when considered separately and positive  when combined}.   Indeed, the second term,  equal to $ \sum_k (\frac{1}{n} \sum_i \mu_{ik} - \frac{1}{K})^2$,  aims at assigning the same probability mass  to the virtual exemplars (and this maximizes  entropy) whilst the third term, equal to $\sum_k \sum_c (f_c(\V_k)-\frac{1}{2})^2$, has also an equivalent behavior w.r.t. its counterpart in Eq.~\ref{eq0},  namely,  it favors virtual exemplars  whose SVM softmax scores are  the closest to $\frac{1}{2}$ (i.e.,  near the SVM decision boundary), and this also maximizes  entropy. This similar behavior is also corroborated through the observed performances on these terms when taken individually\footnote{This equivalence of performances is also due to the fact that no weighting is necessary when these terms are used individually.}; nevertheless,  the joint combination of these  homogeneous terms (through  Eq.~\ref{eq01}) provides a more pronounced  gain compared to their heterogeneous counterparts (in Eq.~\ref{eq0}) as shown later in experiments.

 \subsection{Optimization}

\begin{proposition} 
The optimality conditions of Eqs.~(\ref{eq0}) and (\ref{eq01})  lead  to
 {  
 \begin{equation}\label{eq1}
\begin{array}{lll}
  \mu^{(\tau+1)}& :=&\displaystyle  \textrm{\bf diag} \big(\hat{\mu}^{(\tau+1)} \1_K\big)^{-1} \  \hat{\mu}^{(\tau+1)} ,\\
 \\
 \V^{(\tau+1)} &:= & \hat{\V}^{(\tau+1)} \  \textrm{\bf diag} \big(\1'_n {\mu}^{(\tau)} \big)^{-1},
\end{array}  
\end{equation}}
with $ \hat{\mu}^{(\tau+1)}$, $\hat{\V}^{(\tau+1)}$ being respectively for Eq.~(\ref{eq0}) 
{
\begin{equation}\label{eq2}
\begin{array}{l}
\exp\big(-\frac{1}{\gamma}[\DD(\XX,\V^{(\tau)}) + \frac{\alpha}{n}\1_n  ( \1^\top_K +  \log \frac{1}{n} \1^\top_n \mu^{(\tau)} )]\big)\\
 \\
\XX \ \mu^{(\tau)}+ \beta \sum_{c} \nabla_v f_c(\V^{(\tau)}) \circ  (\1_d \ [\log f_c(\V^{(\tau)})]' + \1_d \1_K'),
\end{array}
\end{equation}}
\noindent and  for Eq.~(\ref{eq01}) 
 {
 \begin{equation}\label{eq2}
\begin{array}{l}
 \exp\big(-\frac{1}{\gamma}[\DD(\XX,\V^{(\tau)}) + \frac{\alpha}{n}\1_n  ( \frac{1}{n} \1^\top_n \mu ^{(\tau)}-\frac{1}{K} \1^\top_K) ]\big)\\ 
 \\
\XX \ \mu^{(\tau)}+ \beta \sum_{c}  \nabla_v f_c(\V^{(\tau)}) \circ  (\1_d \ [ f_c(\V^{(\tau)})- \frac{1}{2} \1_K^\top ]),   
\end{array}  
\end{equation}}
 here  $\circ$ stands for the Hadamard product and  $\textrm{\bf diag}(.)$ maps a vector  to a diagonal matrix.        
\end{proposition} 
In view of space, details of the proof are omitted and  result from the optimality conditions of Eq.~\ref{eq0}'s  and  Eq.~\ref{eq01}'s gradients and lagrangians.   Note that ${\mu}^{(0)}$ and  ${\V}^{(0)}$ are initially set to random values and, in practice, the procedure converges to an optimal solution (denoted as $\tilde{\mu}$, $\tilde{\V}$) in few iterations. This solution defines the most {\it relevant} virtual exemplars of $\D_{t+1}$ (according to criteria in Eqs. \ref{eq0} or~\ref{eq01}) which are used to train the subsequent classifier $f_{t+1}$ (see also algorithm~\ref{alg1}).   Note also that $\alpha$ and $\beta$ are set in order to make the impact of the underlying terms equally proportional; in other words,   inversely proportional to  the number of matrix entries intervening in these terms, and this corresponds to $O({K}^{-1})$  for both $\alpha$ and $\beta$.  Finally,  since $\gamma$ acts as  scaling factor that controls the shape of the exponential function, its setting is iteration-dependent and  proportional to the term inside the  square brackets of that  exponential.  
\begin{algorithm}[!ht]
\footnotesize 
  \KwIn{Images in $\S$, display ${\cal D}_0 \subset {\S}$, budget $T$.}
\KwOut{$\cup_{t=0}^{T-1} (\D_t,\Y_t)$ and $\{f_t\}_{t}$.}
\BlankLine
\For{$t:=0$ {\bf to} $T-1$}{$\Y_t \leftarrow oracle(\D_t)$; \\  
  $f_{t} \leftarrow \arg\min_{f} {Loss}(f,\cup_{k=0}^t (\D_k,\Y_k))$; \\
 $\tau \leftarrow 0$; $\hat{\mu}^{(0)} \leftarrow \textrm{random}$; ${\hat{\V}}^{(0)} \leftarrow \textrm{random}$;\\
Set ${{\mu}}^{(0)}$  and ${{\V}}^{(0)}$  using Eq.~(\ref{eq1});   \\
\BlankLine
 \While{($\footnotesize \|\mu^{(\tau+1)}-\mu^{(\tau)}\|_1 +  \|\V^{(\tau+1)}-\V^{(\tau)}\|_1 \geq\epsilon  \newline \wedge  \tau<\textrm{maxiterations})$}{
 Update ${\mu}^{(\tau+1)}$ and ${\V}^{(\tau+1)}$ using Eq.~(\ref{eq1}); \\  
   $\tau \leftarrow \tau +1$;\\
 }
$\tilde{\mu} \leftarrow \mu^{(\tau)}$;  $\tilde{\V} \leftarrow \V^{(\tau)}$; \\ 
 ${\D_{t+1} \leftarrow \big\{\x_i \in \S\backslash \cup_{k=0}^t \D_k: \x_i\leftarrow\arg\min_\x\|\x-\V_k\|_2^2 \big\}_{k=1}^K}$.
}
\caption{Display selection mechanism}\label{alg1}
\end{algorithm}

\section{Experiments}
Change detection experiments  are conducted on the Jefferson dataset. The latter includes $2,200$ aligned patch pairs  (of $30\times 30$ RGB pixels each) taken from bi-temporal GeoEye-1 satellite images of $2, 400 \times  1, 652$ pixels with  a spatial resolution of 1.65m/pixel.  These images were taken from the area of Jefferson (Alabama) in 2010 and 2011 with many changes (building destruction, etc.)  due to tornadoes as well as no-changes (including irrelevant ones as clouds).   This dataset  includes 2,161 negative  pairs (no/irrelevant changes) and only 39 positive   pairs (relevant changes), so  less than $2\%$ of these data correspond to relevant changes, and this makes  their localization very  challenging.   In   our experiments,  we split the whole dataset evenly; one half to train our display and learning models while the remaining half to measure accuracy.   As  changes/no-changes classes are highly off-balanced, we measure  accuracy using the equal error rate (EER); smaller EER implies better performances.

\subsection{Ablation}
We first study the impact of each term of our objective functions separately,  and then we consider them jointly and all combined.  Note that the regularization term is always kept as it allows to obtain the closed form solutions shown in proposition~1.  Table~\ref{tab1} illustrates  the impact of these terms where EER performances are shown for each configuration using both the early and the surrogate models respectively.  As expected, when using the terms separately, their impact on change detection performances are very similar (and sometimes identical), and this results from their equivalence; put differently, the solution reached by these terms --- even though presented in different ways in their respective objective functions --- have similar behavior when used separately.  However, when jointly combining these terms, their impact is different, and this results from their heterogeneity in Eq. \ref{eq0} and homogeneity in Eq. \ref{eq01} which makes the setting of the underlying mixing hyperparameters in the latter configuration easier and also more effective.  From these results, we  also observe the highest impact of representativity+diversity especially at the earliest iterations of change detection, whilst the impact of ambiguity term raises later in order to locally refine the decision criteria  (i.e., once the modes of data distribution become well explored). These EER performances are shown  for different sampling percentages defined --- at each iteration $t$ --- as $(\sum_{k=0}^{t-1} |\D_k|/(|\I|/2))\times 100$  with   $|\I|=2,200$ and $|\D_k|$ set to $16$. 

 \begin{table}\centering
 \resizebox{0.97\columnwidth}{!}{
 \begin{tabular}{ccc||cccccccccc||c}
 rep  & div & amb & 1 &2 & 3& 4& 5& 6& 7& 8& 9 & 10 & AUC. \\
 \hline
 \hline

        \xmark  &  \xmark &     \cmark    & 47.81    & 27.29  &  11.15  &  7.97 &   8.18  &  7.31  &  7.97  &  7.94 &   7.50  &  7.90 &  14.10\\
         & & &  47.81 &    18.71 &    11.23 &    7.96 &     8.17  &    7.28 &    7.58 &     7.88 &     7.49 &     7.90& 13.21 \\
         \hline 
         \xmark  & \cmark  &       \xmark  & 47.81    & 18.72  &  11.24&     7.97 &   8.18&     7.29&     7.59   &  7.88   &  7.50  &   7.90 & 13.21  \\
         &  &  &  47.81 &   18.71 &   11.23 &    7.96 &    8.17 &    7.28 &    7.58 &    7.88 &    7.49 &    7.90 & 13.21 \\ \hline 
       
         \cmark   &  \xmark  &      \xmark &  47.81   &  35.98 &   16.86 &   6.52 &    4.98&     2.67 &    2.03   &  1.80 &    1.45&     1.30 & 12.14 \\
  & & &  47.81   &  35.98 &   16.86 &   6.52 &    4.98&     2.67 &    2.03   &  1.80 &    1.45&     1.30 & 12.14   \\ \hline  
         \cmark   &      \xmark &   \cmark  &  47.81   &  40.40&   23.86&     9.56&    7.65&     5.75 &    5.47  &   6.12 &    4.40   &  5.72 &  15.67 \\
       & & &    47.81 &   36.75 &   29.26 &     8.61 &    4.27 &    2.37 &    2.34 &    1.68 &    1.45 &    1.27 & 13.59 \\
         \hline 
         \xmark  &  \cmark &   \cmark  &  47.81   & 27.29  &   11.15    & 7.97  &   8.18   &  7.31   &  7.97   &  7.94  &   7.50  &   7.90 & 14.10  \\
      &  & &    47.81 &   28.94 &  12.39 &    9.12 &    7.05 &     6.94 &    7.05&     7.09 &    7.25 &    6.93 & 14.06 \\ \hline 
         \cmark  & \cmark    &     \xmark  & 47.81   &  29.84 &    17.63 &    6.21  &   4.40   & 2.70  &   1.98 &    1.92  &   1.65   &  1.52 & 11.57 \\
         & & &  47.81 &   38.20 &   23.73 &   9.36 &    7.67 &    5.67 &    4.66 &    4.31 &    3.28 &    2.59 & 14.73 \\
\hline
        \cmark   &  \cmark  &  \cmark  &  47.81  &  27.61  &   11.76 &    5.74  &   2.95  &   2.39  &   1.89  &   1.61  &   1.55 &    1.34 &10.47 \\
    & & &   47.81  &   32.56&    \bf9.88&    \bf4.54&    \bf2.71&    \bf2.00&    \bf1.56&    \bf1.21&    \bf1.10&    \bf1.08& \bf10.44\\
\hline \hline 
\multicolumn{3}{c||}{Samp\%}  & 1.45 &2.90 & 4.36& 5.81& 7.27& 8.72& 10.18& 11.63& 13.09 & 14.54 & - 
\end{tabular}}
 \caption{This table shows an ablation study of our display model. Here rep, amb and div stand for representativity, ambiguity and diversity respectively.  For each configuration of ``rep'', ``amb'' and ``div'',  the results are shown in two rows corresponding to the early and the surrogate  display models respectively.   These results are  also shown for different iterations $t=0,\dots,T-1$ (Iter) and the underlying sampling rates (Samp) again defined as $(\sum_{k=0}^{t-1} |\D_k|/(|\I|/2))\times 100$. The AUC (Area Under Curve) corresponds to the average of EERs across iterations.}\label{tab1} 
 \end{table} 
 \subsection{Comparison}\label{compare}
\begin{figure}[tbp]\centering
\includegraphics[angle=90,width=0.7\linewidth]{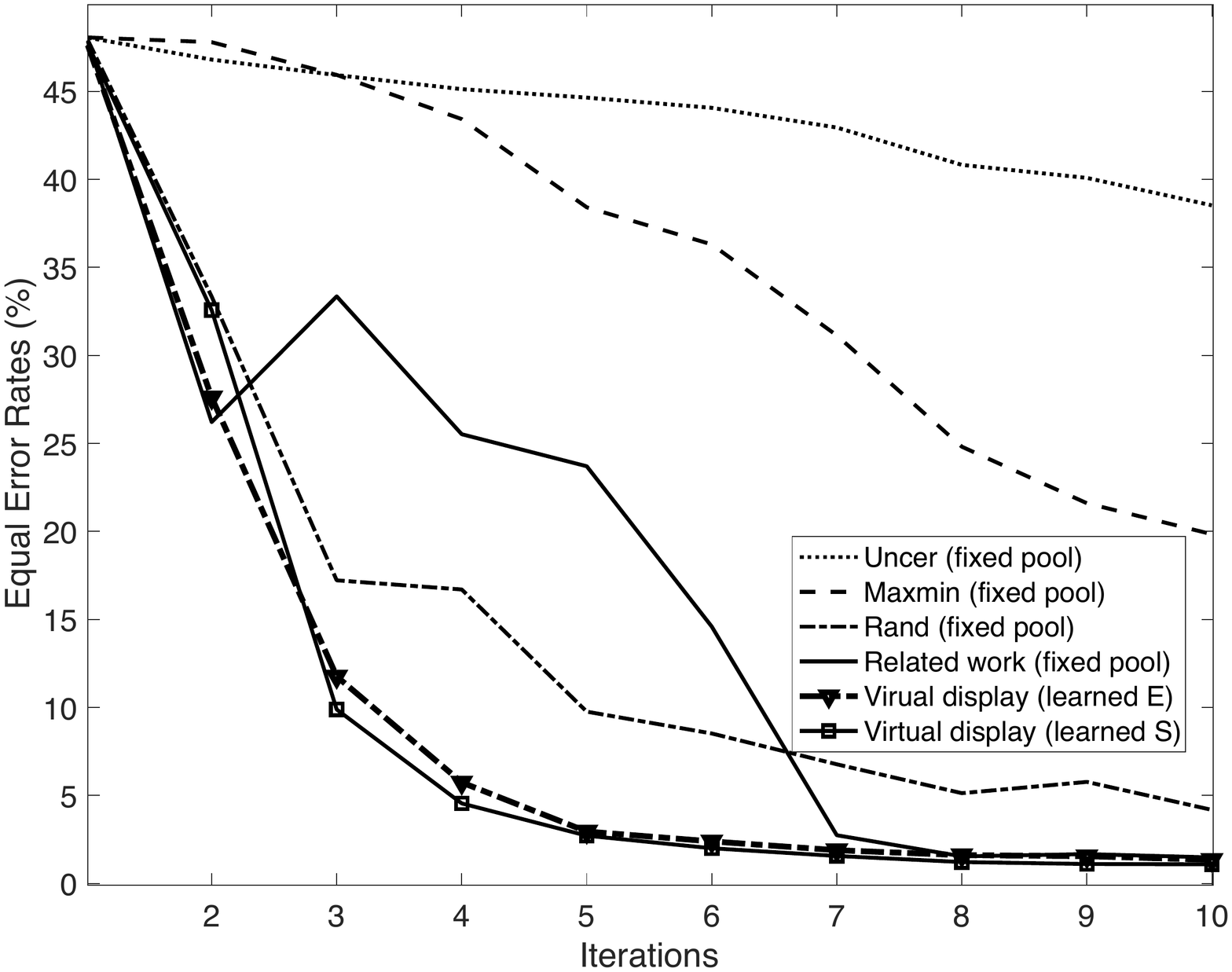}
 \caption{This figure shows a comparison of different sampling strategies w.r.t. different iterations (Iter) and the underlying sampling rates in table~\ref{tab1} (Samp). Here ``Uncer'' and ``Rand'' stand for uncertainty and random sampling respectively while ``learned E'', ``learned S'' correspond to the proposed early and   surrogate display models respectively.  Note that fully-supervised learning achieves an EER of $0.94 \%$.  Related work stands for the method in (Sahbi et al.,  IGARSS 2021); see again section~\ref{compare} for more details.}\label{tab2}\end{figure}

We also compare our display model w.r.t. other display selection  strategies  including  pool-based methods namely  {\it random search, maxmin and uncertainty}. Random consists in sampling displays from the pool of unlabeled data whereas uncertainty  aims at choosing, from the same pool,  the most ambiguous data,  i.e.,  whose SVM scores are the closest to zero.  Maxmin seeks to greedily select the most distinct data in $\D_{t+1}$;  each sample in  $\x_i \in \D_{t+1} \subset \S\backslash \cup_{k=0}^t \D_k$ is chosen by {\it maximizing its minimum distance w.r.t.  $\cup_{k=0}^t \D_k$}.  We further compare our display model against the model in (Sahbi et al.,  IGARSS 2021) which assigns marginal probability measures (instead of conditional ones) to the unlabeled data and selects the data in the subsequent display with the highest probability values.   Finally,  we report performances using a fully-supervised setting that considers the entire training set (whose labels are taken from the ground-truth) in order to learn a single monolithic classifier, and the EER of the latter is considered as a lower bound.  The EER performances of these settings are shown in Figure~\ref{tab2} through different iterations and sampling rates,  as well as the aforementioned display selection strategies.  From these results, most of the comparative methods are powerless to spot the rare class (i.e., changes) sufficiently well.  Indeed, while random and maxmim capture the diversity during the early stages of interactive change detection they are less effective in refining the  learned decision functions whilst uncertainty overcomes this issues,  it lacks diversity.  The display selection strategy in (Sahbi et al.,  IGARSS 2021)  captures diversity and refines better the learned classifiers, however, it suffers from the rigidity of the selected displays (especially at the early iterations) which are taken from a fixed pool of training data.  In contrast,  our proposed display models gather the advantages of the aforementioned methods while they also allow learning more flexible displays that further enhance change detection performances through different iterations.  Besides, the surrogate model leads to an extra gain in performances especially at mid and late iterations of change detection which particularly correspond to highly frugal learning regimes. 
 
\section{Conclusion}

We introduce in this paper a new interactive  satellite image change detection algorithm.  The proposed solution is based on active learning and consists in learning both a classifier and a subset of data (referred to as virtual display) resulting into a more flexible model.   These virtual displays are obtained by maximizing  their  representativity, diversity and also their uncertainty, and this provides a more effective adversarial model that challenges the current trained classifiers leading to more effective subsequent ones.  All these findings are corroborated through extensive experiments conducted on the challenging task of interactive satellite image change detection which show the out-performance of the proposed virtual display models against different baselines and pool-based methods as well as the related work.

\end{document}